\documentclass{article}

\usepackage[final]{tackling_climate_workshop_style}

\usepackage[utf8]{inputenc}
\usepackage[T1]{fontenc}
\usepackage{amsmath, amssymb}
\usepackage{url}
\usepackage{graphicx}
\usepackage{booktabs}
\usepackage{nicefrac}
\usepackage{microtype}     
\usepackage{subcaption} 
\usepackage{multirow}
\usepackage{xcolor,colortbl}
\usepackage{nicefrac}

\usepackage[hidelinks]{hyperref}

\title{A Probabilistic U-Net Approach to Downscaling Climate Simulations}

\author{%
Maryam Alipourhajiagha \quad Pierre-Louis Lemaire \quad Youssef Diouane \quad Julie Carreau \\
Polytechnique Montr\'eal\\
\texttt{\{maryam.alipourhajiagha,pierre-louis.lemaire,youssef.diouane,julie.carreau\}}\\\texttt{@polymtl.ca}\\
}

\begin{document}

\maketitle

\begin{abstract}
Climate models are limited by heavy computational costs, often producing outputs at coarse spatial resolutions, while many climate change impact studies require finer scales. Statistical downscaling bridges this gap, and we adapt the probabilistic U-Net for this task, combining a deterministic U-Net backbone with a variational latent space to capture aleatoric uncertainty. We evaluate four training objectives, afCRPS and WMSE–MS-SSIM with three settings for downscaling precipitation and temperature from $16\times$ coarser resolution. Our main finding is that WMSE–MS-SSIM performs well for extremes under certain settings, whereas afCRPS better captures spatial variability across scales.

\end{abstract}

\section{Introduction}

Climate change is amplifying hazards like heatwaves, extreme weather, and floods, with escalating economic and social impacts~\citep{Quesada-Chacon2023-tx}. Most impact studies require ensembles of high-resolution climate projections, but regional climate models even though they are capable of providing fine-scale variables via dynamic downscaling are computationally expensive, making such ensembles scarce~\citep{climex}.
To circumvent these computational costs, there is growing interest in emulators; statistical models designed to perform downscaling with far less computational power and memory. Many recent emulators leverage advances in deep learning, which offer the flexibility to capture complex spatial patterns~\citep{WuEtal2021,Doury2024,gonzalez2023multi,Rampal2022Interpretable}. 
Generative models may be better suited than purely deterministic models, which are often trained with MSE and tend to produce overly smoothed downscaled fields while missing extreme events~\cite{Sun2024Review}.
Traditional stochastic weather generators struggled to scale over full spatial domains~\cite{AilliotEtal2015}, motivating the use of deep learning approaches such as Generative Adversarial Networks~\citep{Annau2023Hallucinations}, conditional normalizing flows~\citep{WinklerEtal2024}, and diffusion models~\citep{BassettiEtal2024} for climate downscaling.
 
 In this work, we introduce the probabilistic U-Net to climate downscaling. Like the standard U-Net, the probabilistic U-Net was originally developed for medical image segmentation~\citep{kohl2018probabilistic}. 
 Given the widespread use of U-Nets for downscaling~\citep{Sun2024Review}, it is valuable to assess their probabilistic variant in this context.
 In particular, we focus on selecting the most suitable training objective to enhance local-scale variability, avoiding the smoothing effect of MSE and to improve the reproduction of extreme events, which are critical when studying meteorological hazards. We consider daily total precipitation and minimum/maximum temperatures
  downscaled from data at $16\times$ coarser resolution.


\paragraph{Contributions} The key contributions of this work are: (i) the first application of the probabilistic U-Net to climate downscaling, and (ii) an optimized training objective that better captures extremes and fine-scale variability. The implementation is publicly available at \href{https://github.com/MaryamAlipourH/prob-unet-climate-downscaling}{\texttt{github.com/MaryamAlipourH/prob-unet-climate-downscaling}}.

\section{Background}

\paragraph{U-Net Backbone}
We cast downscaling as supervised image-to-image translation using a four-level U-Net patterned on the StyleGAN/EDM backbone~\cite{watt2024generative,Karras2022edm}. The encoder halves spatial resolution four times, doubling the channel count from $64$ to $256$, while the decoder mirrors this process with nearest-neighbour up-sampling followed by $3{\times}3$ convolutions. Each encoder level uses two residual blocks and each decoder level uses three, with skip connections concatenating matching scales. 
Since the U-Net requires matching input–output resolution, we upsample low-resolution fields with nearest-neighbor interpolation to avoid smoothing and artifacts. 
The network predicts the residual between this interpolated field and the true high-resolution target field.

\paragraph{Probabilistic U-Net}
A generative model is obtained by wrapping the deterministic U-Net backbone within the Probabilistic U-Net framework~\citep{kohl2018probabilistic}. A prior network produces $P(z|X)$ from the input alone, while a posterior network produces $Q(z| X,Y)$ when the high-resolution target is available; both distributions are axis-aligned Gaussians. During training, we draw $z \sim Q(z|X,Y)$, broadcast it to a feature map, concatenate it to the final U-Net activations, and pass the result through three $1{\times}1$ convolutions to obtain the prediction $\hat Y$. The loss
\begin{equation}
\mathcal{L} = \mathrm{CE}(Y,\hat Y)
            + \gamma\,\mathrm{KL}\!\bigl(Q(z|X,Y)\,\|\,P(z|X)\bigr)
\label{eq:probunet-loss}
\end{equation}
follows Eq.~(4) of \citep{kohl2018probabilistic}, where CE denotes the cross-entropy loss used for the segmentation task, and KL denotes the Kullback–Leibler divergence; the weight $\gamma$ is adapted after a short warm-up phase. 
At inference time, we sample latent vectors from $P(z|X)$, yielding an ensemble of high-resolution realisations that satisfy the learned distribution. The probabilistic U-Net architecture for statistical downscaling is shown in Fig.~\ref{fig:probunet-arch} (see Appendix). 


\paragraph{Training Objectives} 
In the context of downscaling, the CE loss in \eqref{eq:probunet-loss} is not well suited. Although MSE is a straightforward alternative, it is known to fail to capture extreme values. Furthermore, because the model is generative, a loss that promotes ensemble diversity is preferable. For these reasons, we evaluate two alternative losses. The first, termed WMSE-MS-SSIM, is a weighted loss designed to better capture heavy rainfall events~\cite{HessBoers2022}:

\begin{equation}
L_{\lambda}(Y, \hat Y) = \frac{\lambda}{N} \sum_{i=1}^N w(Y_i) (Y_i - \hat Y_i)^2 + (1-\lambda) ({\rm MS-SSIM}(Y, \hat Y)), 
\label{eq:wmse:msssim}
\end{equation}
where $w(Y_i) = \min\left\{\alpha e^{\beta Y_i},1\right\}$,  MS-SSIM is the so-called multi-scale structural similarity measure, and $\lambda$, $\alpha$ and $\beta$ are hyperparameters. 
The second loss function, called almost fair CRPS (afCRPS), was designed to train a generative model for weather forecasting~\citep{LangEtal2024}:

\begin{equation}
\mathrm{afCRPS}_{\eta}\left(\{\hat Y_{i1}, \dots, \hat Y_{iM} \}, Y_i\right) =
\frac{1}{M} \sum_{j=1}^M |\hat Y_{ij} - Y_i|
- \frac{1 - \epsilon}{M (M - 1)}
\sum_{1 \leq j < k \leq M} |\hat Y_{ij} - \hat Y_{ik}|,
\label{eq:afcrps-stable}
\end{equation}
with $M$ the number of simulations generated by the model for a given sample $i$, $\epsilon= \frac{1 - \eta}{M}$, and $\eta$ is an hyperparameter.  
We assess four losses for training the Probabilistic U-Net:  WMSE–MS-SSIM \eqref{eq:wmse:msssim}, with $\alpha=0.007$ and $\beta=0.048$ fixed at their tuned values, and (i) $\lambda$ set to 1 (WMSE only), (ii) 0 (MS-SSIM only), or (iii) the tuned value $0.158$ \cite{HessBoers2022}; and (iv) afCRPS \eqref{eq:afcrps-stable}, with $\eta=0.95$ as in \citep{LangEtal2024}.

\section{Experiments}

\paragraph{ClimEx Daily Meteorological Data:}
We use one member of the ClimEx ensemble of dynamically downscaled simulations ~\citep{climex} over southern Quebec and the Canadian Maritimes at 0.11° ($\approx12$km) resolution, considering total precipitation (in mm)  and minimum/maximum temperatures (in $^\circ$C). The high-resolution domain has $128 \times 128$ grid cells; low-resolution data are obtained by averaging $16 \times 16$ blocks, yielding an $8 \times 8$ grid. Training used 1960–1990, validation 1990–1997, and testing 1998–2005, avoiding the period when the RCP8.5 scenario begins. For the estimation of return levels, we extend the test set to 30 years in order to capture a complete cycle of climate variability.

\paragraph{Experimental Setup:}
Two physical constraints are enforced through re-parametrization: precipitation is kept non-negative using the softplus function $\log(1+e^{x+c})$ ($c=10^{-7}$), and $T_{\max} \geq T_{\min}$ is ensured by applying it to $T_{\max} - T_{\min}$.
%
All performance metrics and losses are computed after converting predictions back to physical units.
We trained the model for 10 epochs with batch size of 32. The latent space was set to a dimension of 16. During training, the Kullback–Leibler divergence term in \eqref{eq:probunet-loss} was gradually scaled to control its relative contribution to the total loss.

\subsection{Qualitative Evaluation}
\paragraph{Return Levels}
Return level curves are often used by practitioners to assess the probability of extreme events, and they can therefore be used to evaluate the ability of downscaling methods to reproduce such events. For a given grid cell, the $T$-year return level can be defined as the quantile of the distribution of annual maxima associated with exceedance probability $1/T$~\cite{coles2001introduction}. We construct return level curves with 95\% confidence bands using the ground truth (i.e., the target test data) by fitting a Generalized Extreme Value distribution to the annual maxima at each grid cell and applying a parametric bootstrap to obtain the confidence bands. Empirical return levels from 5 predictions of the probabilistic U-Net over the test period are then superimposed for comparison, and the match is considered good if the empirical return levels lie within the confidence bands for at least 95\% of points.
Fig.~\ref{fig:plots} (precipitation) and Fig.~\ref{fig:rl_temperature} in the Appendix (maximum/minimum temperatures) show the results for two grid cells. Among the three WMSE–MS-SSIM variants, the tuned setting ($\lambda = 0.158$) performs best, while the afCRPS variant tends to overshoot extremes.

\begin{figure}[ht]
   \includegraphics[width=0.48\linewidth]{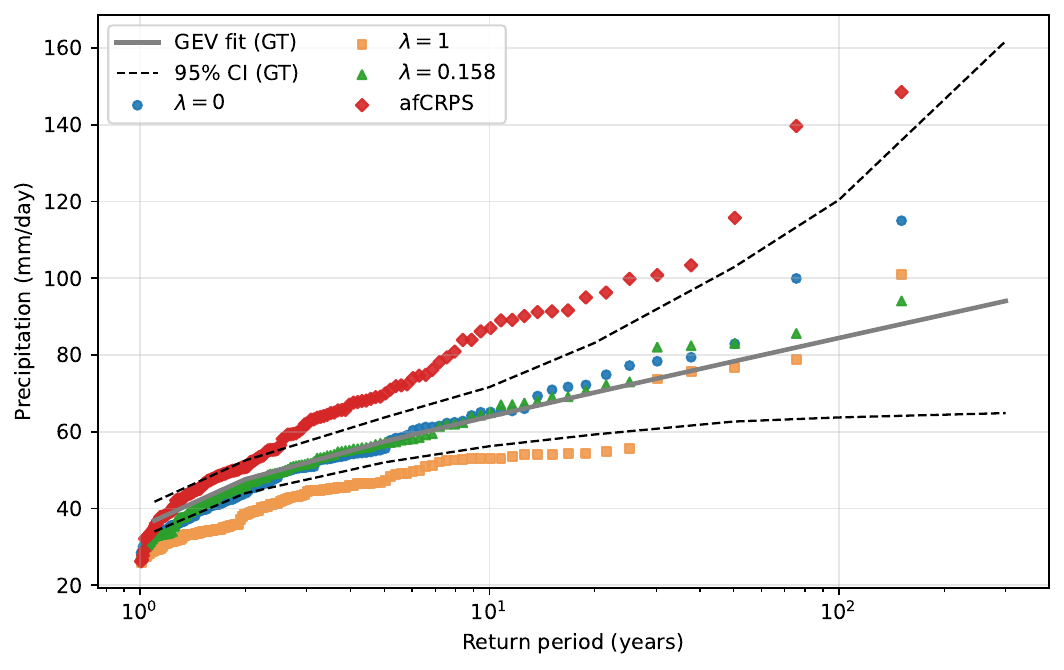}
\includegraphics[width=0.48\linewidth]{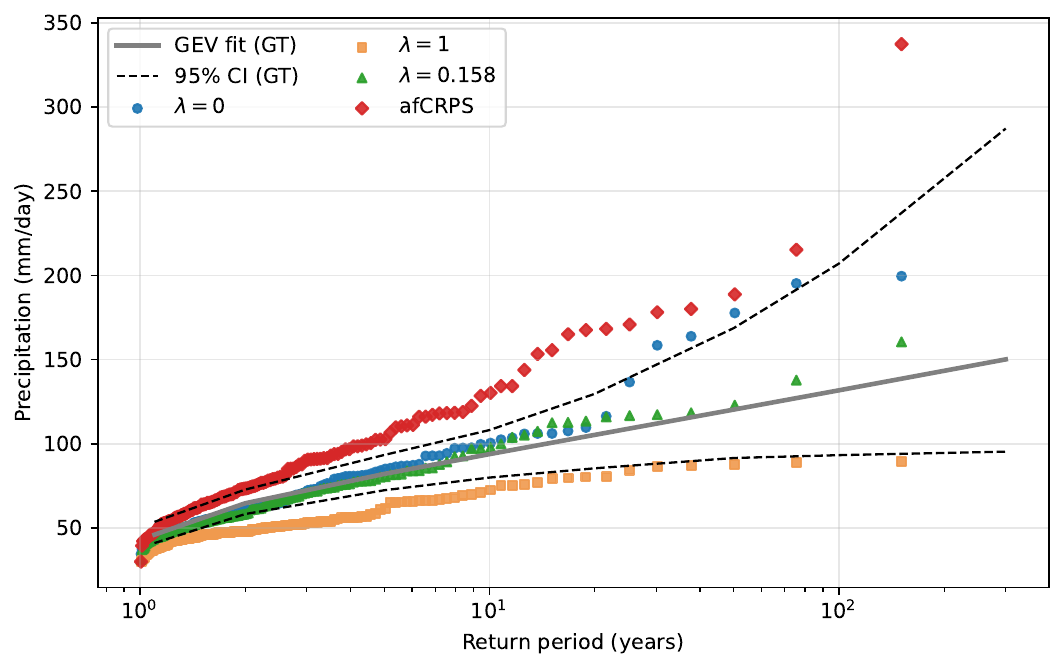}
    \caption{Precipitation return levels for four training objective variants at two grid cells.}
    \label{fig:plots}
\end{figure}

\paragraph{Log-Frequency Histograms}
Because return level curves are pixelwise, we further assess distributional fidelity using log-frequency histograms across all pixels. Fig.~\ref{fig:plots} (left) shows the results for precipitation. The $\lambda=1$ variant (WMSE) substantially underestimates high-intensity precipitation, failing to capture extremes. In contrast, $\lambda=0$ (MS-SSIM) and $\lambda=0.158$ better reproduce the observed tail behavior, closely matching the ground truth. The afCRPS variant, however, tends to overestimate extreme events, consistent with its performance in the return level analysis.
Fig.~\ref{fig:hist:temp} in the Appendix shows that minimum and maximum temperature histograms align well with the ground truth across all training objective variants, with only minor deviations at the extremes. This suggests that temperature distributions are relatively insensitive to the choice of training objective, whereas precipitation extremes remain challenging.
\begin{figure}[ht]

\includegraphics[width=0.48\linewidth]{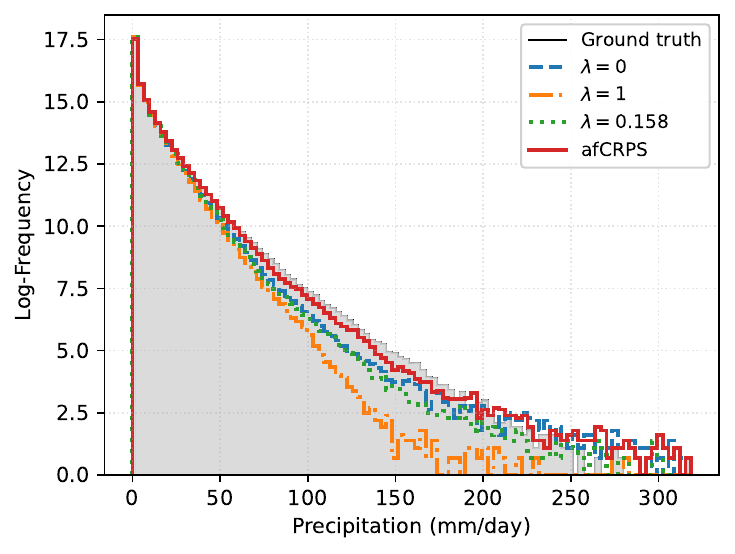}
\includegraphics[width=0.48\linewidth]{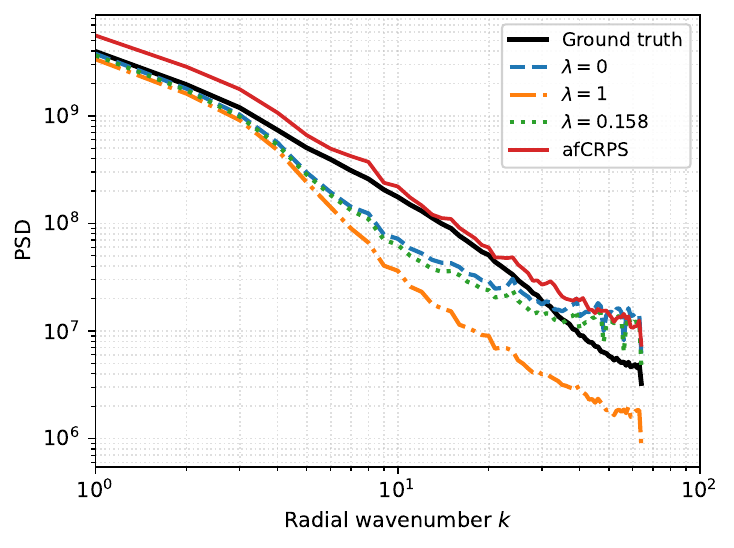}
    \caption{
    Ground truth versus four training objectives for precipitation: log-frequency histograms (left) and power spectral density (right).
    }
    \label{fig:plots}
\end{figure}
\paragraph{Power Spectral Density (PSD)} We evaluate spatial scale fidelity using the azimuthally averaged PSD, which quantifies the distribution of variance across spatial scales. We focus on fine-scale variability, which statistical downscaling often fails to capture.

\begin{equation}
P(k) = \left\langle \, \big| \hat{X}(k) \big|^2 \, \right\rangle \;\;\; \text{with } |k| = k,
\end{equation}
The PSD quantifies how variance is distributed across spatial scales, with low $k$ representing synoptic patterns and high $k$ fine-scale variability.

Fig.~\ref{fig:plots} (right) and Fig.~\ref{fig:psd_temp} (in the Appendix) show PSDs for precipitation and minimum/maximum temperatures, respectively. For precipitation, the WMSE variant ($\lambda=1$) exhibits spectral smoothing and underestimates variance at higher radial wavenumbers. The MS-SSIM variant ($\lambda=0$) better recovers small scales, while the afCRPS model provides the closest match to the observed spectrum across scales. In contrast, temperature fields are well reproduced spectrally by all variants, with only minor deviations at high wavenumbers.
%

\subsection{Quantitative Evaluation}
Table~\ref{tab:quant} mirrors the qualitative trends: none of the variants dominates across all metrics, but different losses emphasize different aspects of skill. Notably, afCRPS improves CRPS overall and excels on temperature MAE while nearest-neighbor interpolation serves as a baseline reference.

\begin{table}[ht]
  \caption{CRPS and MAE for the probabilistic U-Net trained with four different loss functions.}
  \label{tab:quant}
  \centering
  \resizebox{\linewidth}{!}{
  \begin{tabular}{lcccccc}
    \toprule
    \multirow{2}{*}{Loss fn} & \multicolumn{3}{c}{CRPS} & \multicolumn{3}{c}{MAE} \\
    \cmidrule(lr){2-4}\cmidrule(lr){5-7}
     & pr (mm/day) & $T_{\min}$ ($^{\circ}$C) & $T_{\max}$ ($^{\circ}$C) & pr (mm/day) & $T_{\min}$ ($^{\circ}$C) & $T_{\max}$ ($^{\circ}$C) \\
    \midrule
    afCRPS          & \textbf{0.94 $\pm$ 0.74} & \textbf{0.68 $\pm$ 0.20} & 0.62 $\pm$ 0.12 & 1.35 $\pm$ 1.09 & \textbf{0.90 $\pm$ 0.28} & 0.75 $\pm$ 0.17 \\
    $\lambda=0$     & 1.07 $\pm$ 0.85 & 0.86 $\pm$ 0.28 & 0.68 $\pm$ 0.14 & 1.29 $\pm$ 1.00 & 1.06 $\pm$ 0.31 & 0.88 $\pm$ 0.16 \\
    $\lambda=1$     & 1.13 $\pm$ 0.90 & 0.78 $\pm$ 0.26 & \textbf{0.59 $\pm$ 0.14} & \textbf{1.19 $\pm$ 0.94} & 0.94 $\pm$ 0.27 & \textbf{0.74 $\pm$ 0.15} \\
    $\lambda=0.158$ & 1.06 $\pm$ 0.84 & 0.85 $\pm$ 0.27 & 0.66 $\pm$ 0.14 & 1.27 $\pm$ 0.98 & 1.05 $\pm$ 0.30 & 0.85 $\pm$ 0.16 \\
    NN              & -- & -- & -- & 1.51 $\pm$ 1.14 & 1.76 $\pm$ 0.60 & 1.30 $\pm$ 0.30 \\
    \bottomrule
  \end{tabular}}
\end{table}

\section{Conclusion}
We demonstrate the successful application of the probabilistic U-Net to climate downscaling, offering potential advantages over its deterministic counterpart, including uncertainty quantification and latent-space interpretability.

Our experiments highlight that no single loss function fully addresses the dual challenges of capturing extremes and fine-scale variability in statistical downscaling. For extremes, MS-SSIM ($\lambda=0$) proved most effective, closely reproducing observed return levels and tail behavior. For small-scale variability, afCRPS provided the best match to the observed spectra, though it tended to overestimate extremes. This trade-off suggests that combining afCRPS with MS-SSIM may offer a more balanced solution. Quantitative metrics further reinforced this complementarity: afCRPS achieved the lowest CRPS and strong overall accuracy, while MS-SSIM variants better represented precipitation extremes despite higher aggregate error.

For impact studies, especially in hydrology, both local-scale variability and extreme events drive risk assessment. Reliable downscaling requires not only accurate averages but also realistic extremes and spatial detail. By exploring loss functions that balance these objectives, probabilistic deep learning models like the Probabilistic U-Net can become valuable tools for climate change impact assessments.



\bibliographystyle{unsrt}
\bibliography{references}

\appendix
\section{Supplementary Material}
\begin{figure}[ht]
    \centering
    \begin{subfigure}{0.48\linewidth}
        \centering
        \includegraphics[width=\linewidth]{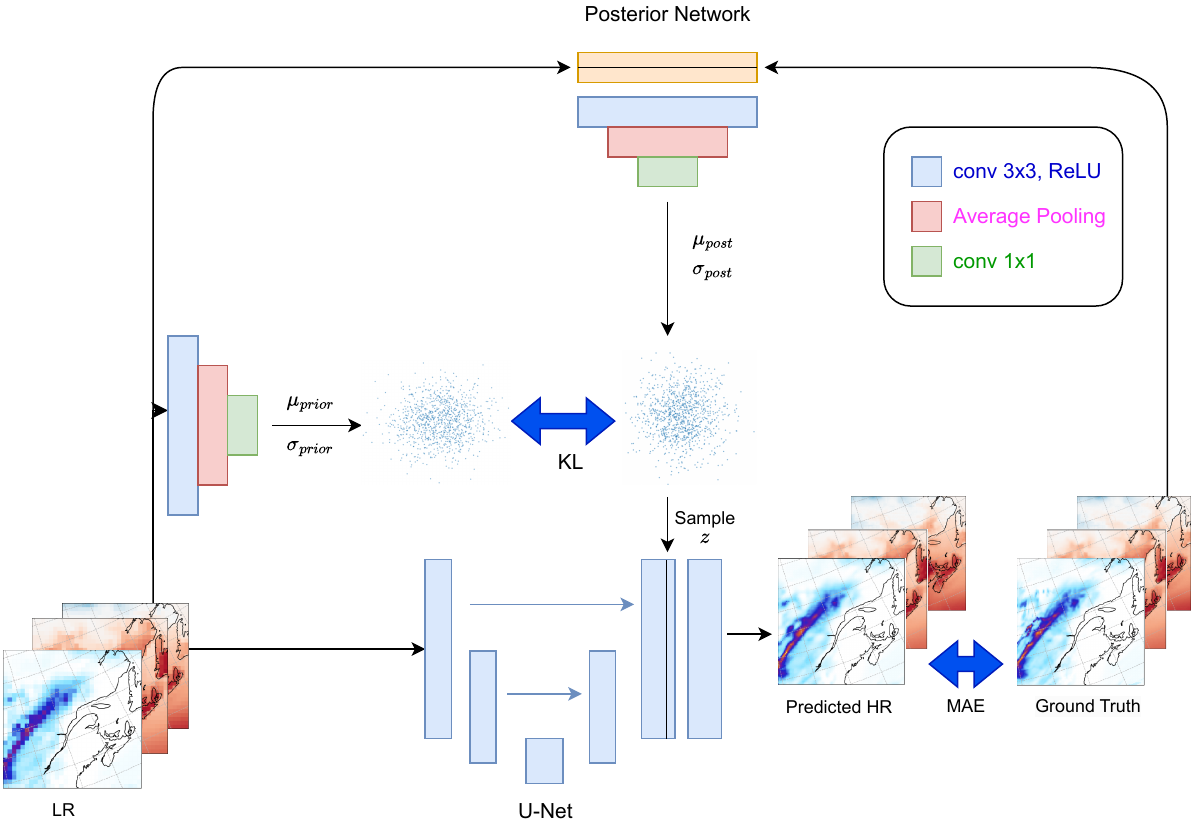}
        \caption{The training phase.}
        \label{fig:probunet-training}
    \end{subfigure}
    \hfill
    \begin{subfigure}{0.48\linewidth}
        \centering
        \includegraphics[width=\linewidth]{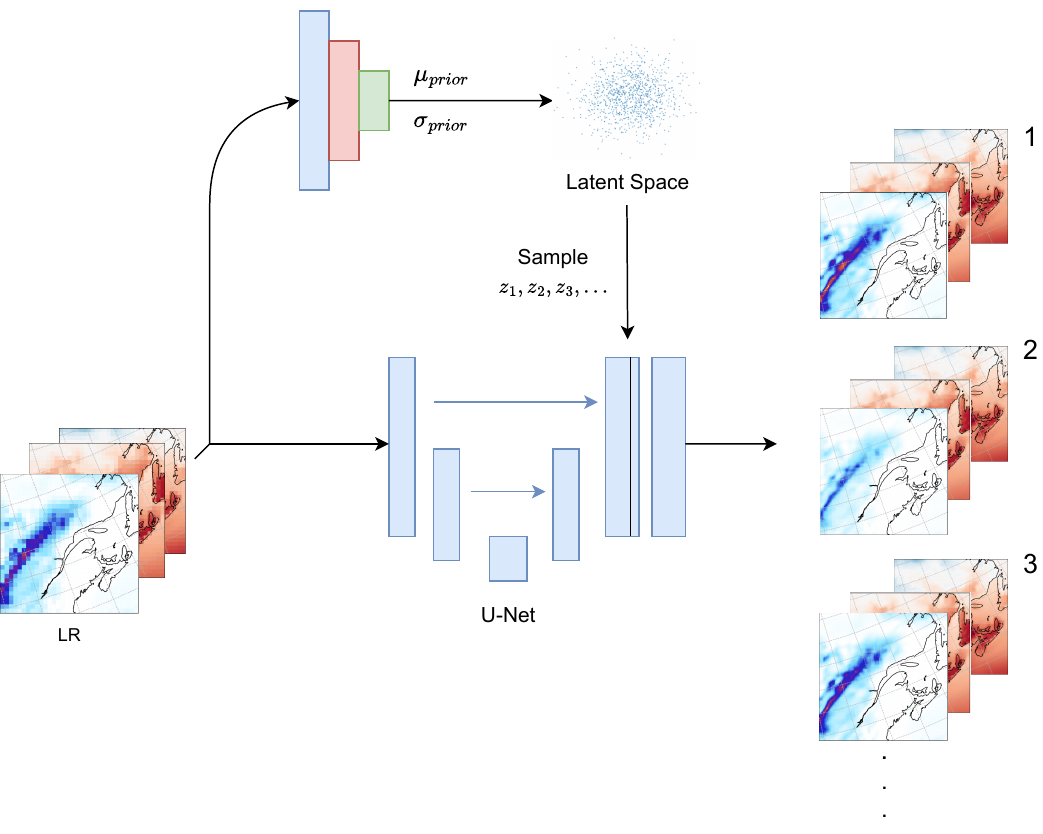}
        \caption{The inference phase.}
        \label{fig:probunet-inference}
    \end{subfigure}
    \caption{Probabilistic U-Net architecture for statistical downscaling, showing the prior and posterior networks, the U-Net backbone, and the latent variable fusion during \textbf{(a)} training and \textbf{(b)} inference.}
    \label{fig:probunet-arch}
\end{figure}

\begin{figure}[htbp]
    \centering
    \includegraphics[width=1.1\textwidth]{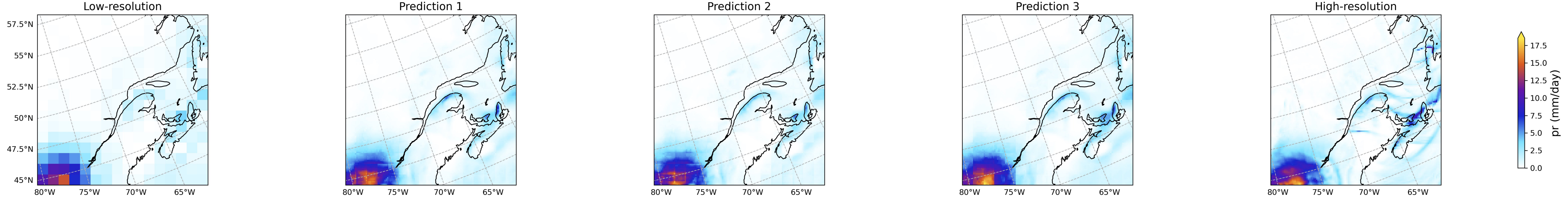}
    \caption{From left to right: the coarse-resolution input, three sampled high-resolution realizations from the model (out of an arbitrarily large ensemble), and the ground-truth high-resolution field. This figure illustrates how the probabilistic U-Net generates diverse yet physically consistent realizations. While the large-scale precipitation pattern is reproduced across all predictions, variability appears in regions of higher intensity, reflecting the model’s stochastic sampling of fine-scale structures. This ensemble spread is precisely what enables the model to represent uncertainty in extremes that a deterministic baseline would smooth out. }
    \label{fig:placeholder}
\end{figure}

\begin{figure}[ht]
    \centering
        \includegraphics[width=.48\linewidth]{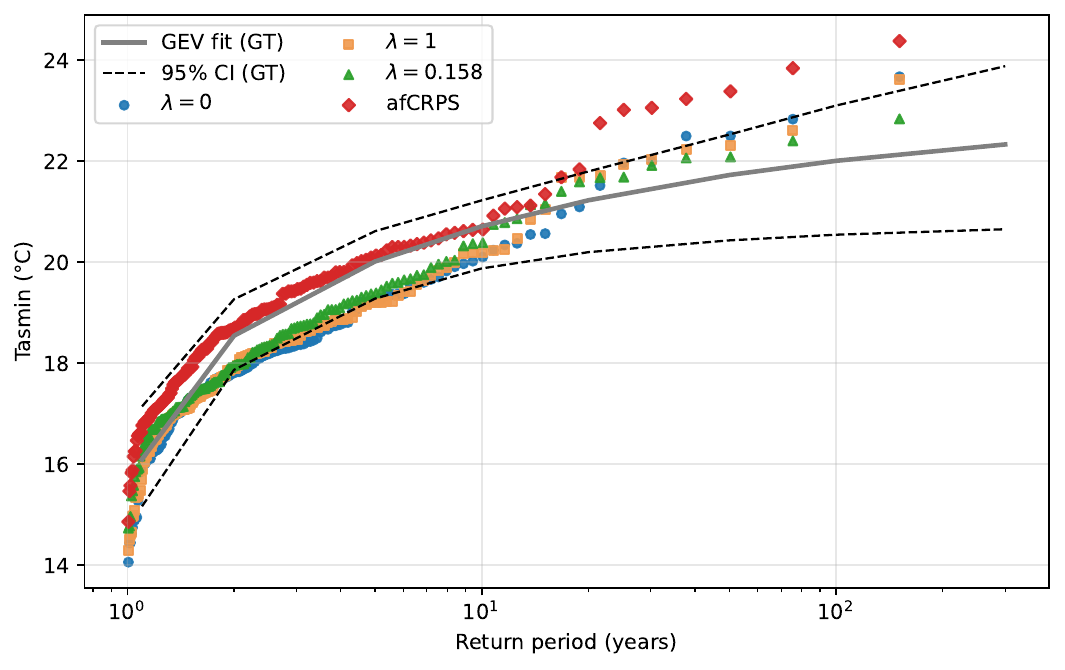}
        \includegraphics[width=.48\linewidth]{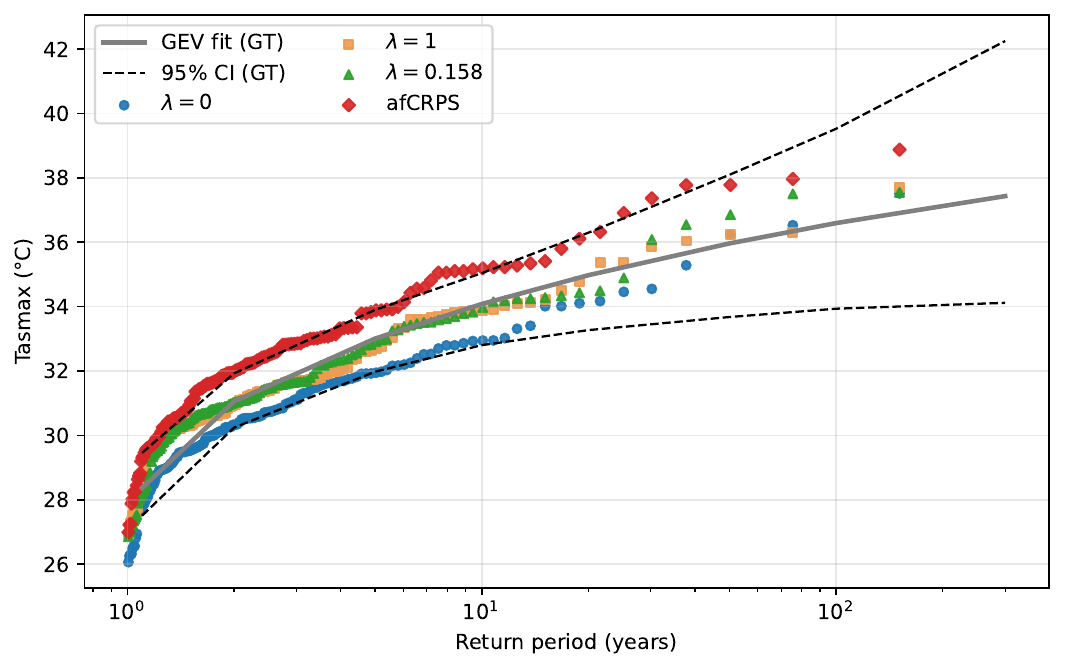}
    \caption{Minimum (left panel) and maximum (right panel) temperature return levels for four training objective variants at two grid cells.}
    \label{fig:rl_temperature}
\end{figure}

\begin{figure}[htpb]
    \centering
    \includegraphics[width=1.0\linewidth]{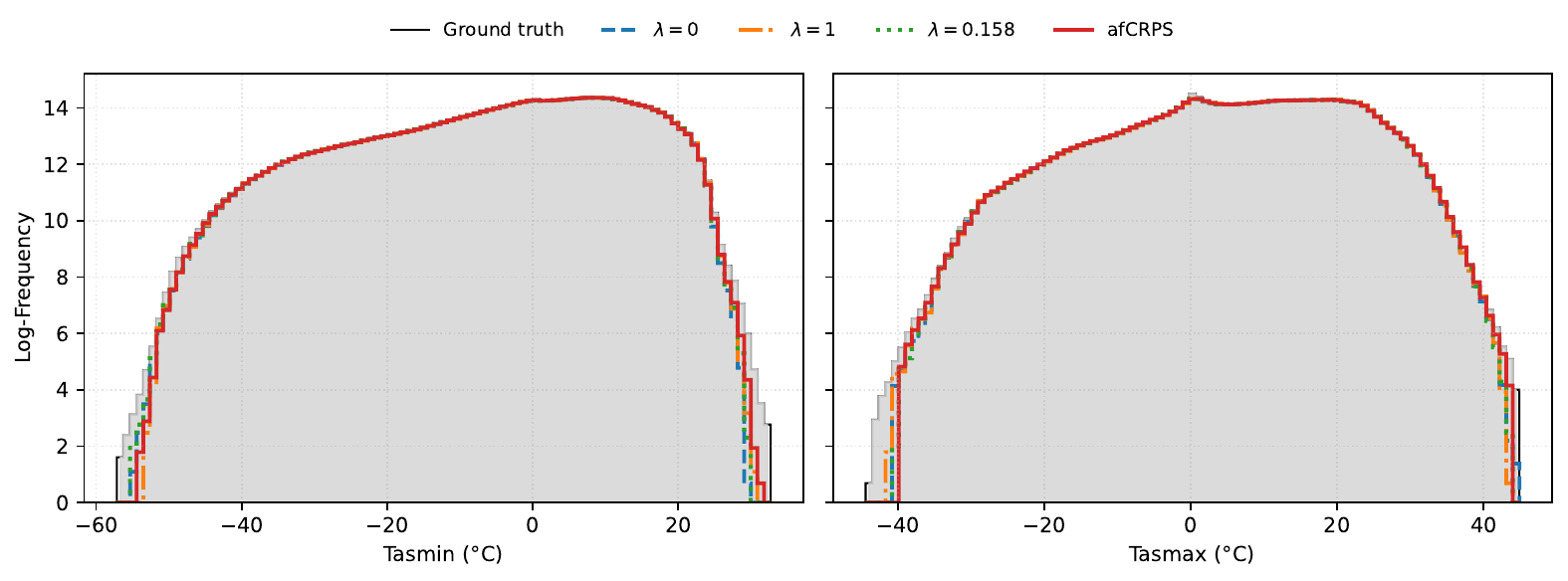}
    \caption{Log-frequency histograms of ground truth versus four training objectives for maximum (left) and minimum (right) temperature.}
    \label{fig:hist:temp}
\end{figure}

\begin{figure}[ht]
    \centering
        \includegraphics[width=.48\linewidth]{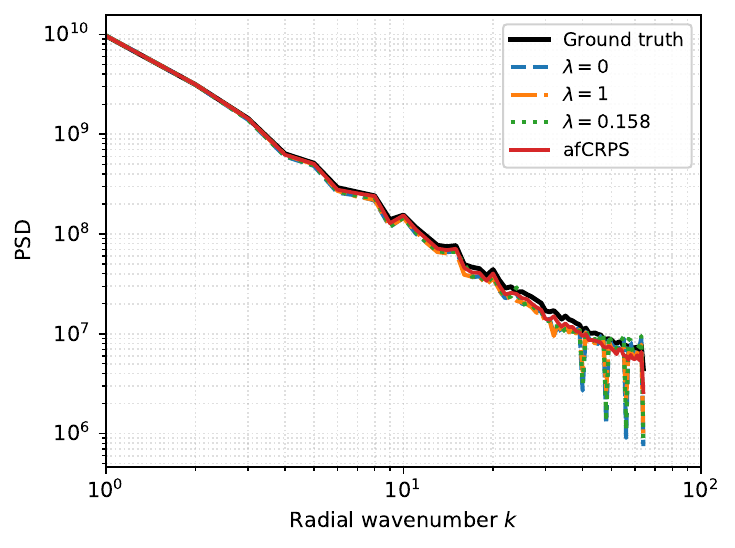}
        \includegraphics[width=.48\linewidth]{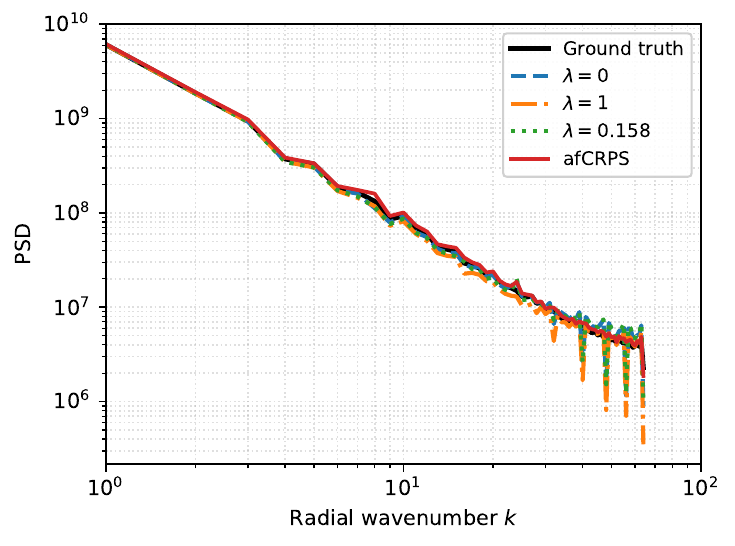}
    \caption{Power spectral density of ground truth versus four training objectives for maximum (left) and minimum (right) temperature.}
    \label{fig:psd_temp}
\end{figure}
\small

\end{document}